\documentclass[manuscript,screen]{acmart}

\AtBeginDocument{%
  }

\setcopyright{acmlicensed}
\copyrightyear{2018}
\acmYear{2018}
\acmDOI{XXXXXXX.XXXXXXX}

\acmISBN{978-1-4503-XXXX-X/18/06}

\usepackage{algorithm}
\usepackage{amsmath}
\usepackage{array}
\usepackage{caption}
\usepackage{float}
\usepackage{graphicx}
\usepackage{listings}
\usepackage{longtable}
\usepackage{mathrsfs}
\usepackage{multirow}
\usepackage{multicol}
\usepackage{pifont}
\usepackage{siunitx}
\usepackage{subfigure}
\usepackage{subfloat}
\usepackage{xcolor}

\usepackage{url}

\begin{document}

\title{A Survey of Research in Large Language Models for Electronic Design Automation}


\author{Jingyu Pan}
\email{jingyu.pan@duke.edu}
\orcid{0000-0002-7187-5205}
\author{Guanglei Zhou}
\email{guanglei.zhou@duke.edu}
\orcid{0000-0002-6840-7160}
\author{Chen-Chia Chang}
\email{chenchia.chang@duke.edu}
\orcid{0000-0003-3115-0733}
\author{Isaac Jacobson}
\email{isaac.jacobson@duke.edu}
\orcid{0009-0000-3672-7058}
\affiliation{
  \institution{Duke University}
  \city{Durham}
  \state{North Carolina}
  \country{USA}
}



\author{Jiang Hu}
\email{jianghu@tamu.edu}
\orcid{0000-0003-1157-7799}
\affiliation{
  \institution{Texas A\&M University}
  \city{College Station}
  \state{Texas}
  \country{USA}
}

\author{Yiran Chen}
\email{yiran.chen@duke.edu}
\orcid{0000-0002-1486-8412}
\affiliation{
  \institution{Duke University}
  \city{Durham}
  \state{North Carolina}
  \country{USA}
}

\renewcommand{\shortauthors}{Trovato et al.}


\begin{abstract}
Within the rapidly evolving domain of Electronic Design Automation (EDA), Large Language Models (LLMs) have emerged as transformative technologies, offering unprecedented capabilities for optimizing and automating various aspects of electronic design.
This survey provides a comprehensive exploration of LLM applications in EDA, focusing on advancements in model architectures, the implications of varying model sizes, and innovative customization techniques that enable tailored analytical insights.
By examining the intersection of LLM capabilities and EDA requirements, the paper highlights the significant impact these models have on extracting nuanced understandings from complex datasets.
Furthermore, it addresses the challenges and opportunities in integrating LLMs into EDA workflows, paving the way for future research and application in this dynamic field.
Through this detailed analysis, the survey aims to offer valuable insights to professionals in the EDA industry, AI researchers, and anyone interested in the convergence of advanced AI technologies and electronic design.
\end{abstract}


\keywords{Do, Not, Us, This, Code, Put, the, Correct, Terms, for,
  Your, Paper}

\received{6 September 2024}
\received[revised]{23 December 2024}
\received[accepted]{16 January 2025}

\maketitle

\section{Introduction}

\begin{figure*}[tb]
    \centering
    \includegraphics[width=0.95\textwidth]{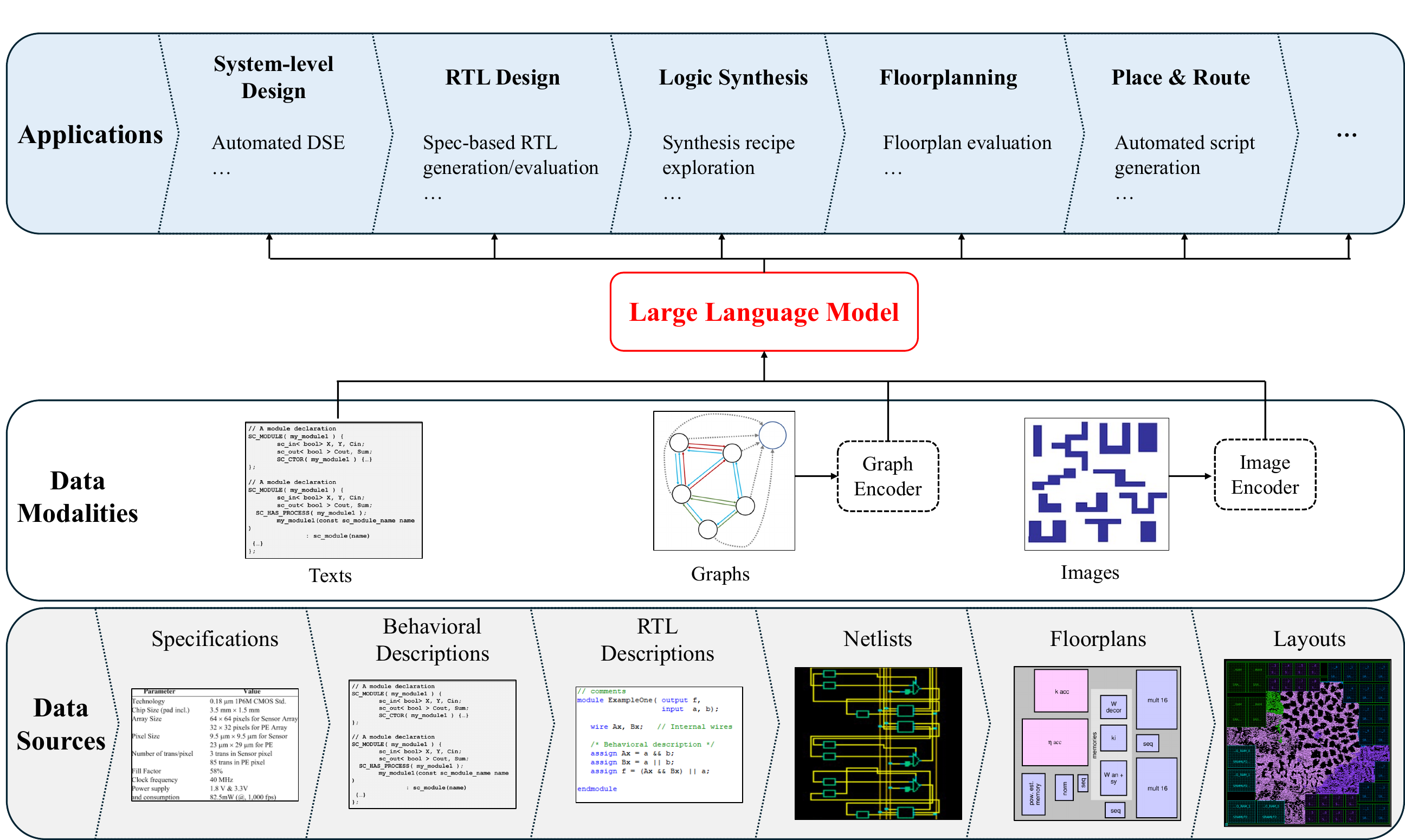}
    \captionsetup{justification=centering}
    \caption{Overview of Large Language Models for EDA applications. Data sources comprise circuit representations from different design stages. Earlier stages typically uses textual data for semantic representations; Later stages involves data of a wider range of modalities, including graphs and images, for detail representations. With proper encoders applied, the circuit data can be processed by the LLMs for novel applications in each design stage.}
    \label{fig:overview}
\end{figure*}

In recent years, Large Language Models (LLMs) have risen prominently in the field of machine learning (ML).
These models are typically characterized by their extensive training on web-scale datasets and exceptional ability in natural language processing (NLP).
In NLP, models such as GPT-3\cite{brown2020language} and its successors\cite{achiam2023gpt} have significantly advanced the capabilities of natural language generation, enabling applications ranging from sophisticated conversational agents to automated content creation.
Simultaneously, the emergence of instruction-tuned models like InstructGPT\cite{ouyang2022training} and RLHF-trained models\cite{li2023reinforcement,christiano2017deep} have improved the alignment of generated content with user intentions and ethical guidelines.
Moreover, the integration of multimodal capabilities, as seen in models like DALL-E\cite{ramesh2021zero} and CLIP\cite{radford2021learning}, has expanded the utility of LLMs beyond text, allowing for seamless processing and generation of text and images, thereby pushing the boundaries of artificial general intelligence.

Despite that LLMs have found downstream applications across various domains, especially those with rich well-structured data, the field of circuit design still poses significant challenges to researchers in both LLM and electronic design automation (EDA).
Despite the fact that designing a circuit is usually a highly iterative process, designers frequently face the challenge of recreating or redesigning circuits from scratch, driven by the subtle yet critical nuances required to meet ambitious performance, power, and area (PPA) objectives.
This repetitive process highlights the need for a learning solution that can effectively draw from historical successes and failures, leveraging the power of LLMs to streamline and optimize the design iterations.


This survey aims to explore the multifaceted applications of LLMs in the EDA domain.
It delves into how these models are revolutionizing the way electronic components are designed, optimized, and brought to market.
From automating mundane tasks to facilitating advanced design simulations, LLMs are redefining the boundaries of what is possible in electronic design.
Through this exploration, the survey aims to offer valuable insights to professionals in the EDA industry, AI researchers, and anyone interested in the intersection of advanced AI technologies and electronic design.
As we stand at the cusp of a new era in electronic design automation, the advent of LLMs marks a significant shift, promising a surge in both efficiency and innovation, while fundamentally changing the way electronic systems are designed and manufactured.

This survey aims to provide a comprehensive understanding of the current state and future potential of LLM applications in EDA.
In the following sections, we will delve into the background on LLM advancements, explore recent trends in their application to EDA, and examine specific stages of electronic design where LLMs are making an impact, including system-level design, RTL design, logic synthesis, and physical design.
Additionally, we will discuss the methodology behind the selection and customization of LLMs, feature representation, and the current academic infrastructure supporting this research.
Finally, we will address the application bottlenecks and future outlook for LLMs in EDA, highlighting both the opportunities and challenges in this emerging field.

\section{Background on LLM}


Recent advances in natural language processing have led to the development of powerful language models such as the GPT series~\cite{radford2018improving}, including large language models (LLM) such as ChatGPT and GPT-4~\cite{achiam2023gpt}.
These models are pre-trained on vast amounts of text data and have demonstrated exceptional performance in a wide range of NLP tasks, including language translation, text summarization, and question-answering.
In particular, the ChatGPT model has demonstrated its potential in various ﬁelds, including education, healthcare, reasoning, text generation, human-machine interaction, and scientiﬁc research.

A key milestone of LLM development is InstructGPT~\cite{ouyang2022training}, a framework that allows for instruction ﬁne-tuning of a pre-trained language model based on Reinforcement Learning from Human Feedback (RLHF)~\cite{ouyang2022training,christiano2017deep}.
This framework enables an LLM to adapt to a wide range of NLP tasks, making it highly versatile and flexible by leveraging human feedback. RLHF enables the model to align with human preferences and human values, which significantly improves from large language models that are solely trained text corpora through unsupervised pre-training. ChatGPT is a successor to InstructGPT. Since its release in December 2022, ChatGPT has been equipped with these advanced developments, leading to impressive performance in various downstream NLP tasks such as reasoning and generalized text generation. These unprecedented NLP capabilities spur applications in diverse domains such as education, healthcare, human-machine interaction, medicine, and scientific research. ChatGPT has received widespread attention and interest, leading to an increasing number of applications and research that harness its extraordinary potential. The open release of the multimodal GPT-4 model further expands the horizon of large language models and empowers exciting developments that involve diverse data beyond text. In response to ChatGPT, other competitors in the market have released powerful proprietary LLMs such as Perplexity's Claude3 \cite{anthropic2024claude3}. The Claude3 family of multimodal models includes three models of different sizes, with Claude3 Opus being the largest and most powerful.

In addition to proprietary models such as GPT-4 and Claude, the open source community has also made significant strides in developing LLMs. LLaMA3 \cite{grattafiori2024llama3herdmodels} and its predecessors LLaMA \cite{touvron2023llama} and LLaMA2 \cite{touvron2023llama2} are a series of LLMs designed by Meta and freely available.
LLaMA was introduced to support researchers and engineers in advancing the field with models that offer state-of-the-art performance while being smaller and more computationally efficient than their competitors. LLaMA 2 and 3, improved on this foundation with increased performance and usability, further democratizing LLM research.
More recent releases from the LLaMA3 family included support for multiple modalities in LLaMA3.2 \cite{meta2024llama3.2}.
CodeGen\cite{nijkamp2022codegen} is an open source model designed specifically for code generation. It is trained on a diverse set of code repositories and designed to assist with various programming tasks. It provides significant improvements in tasks related to code completion, summarization, and translation between different programming languages.
Mistral 7B\cite{jiang2023mistral} is a powerful open-source LLM developed by Mistral AI, featuring 7 billion parameters. It uses a refined attention mechanism and grouped-query attention (GQA) to enhance inference speed and reduce memory usage. It has shown competitive performance in a variety of NLP tasks while being compact and efficient, making it an attractive choice for many applications. Mistral AI has recently released a 12 billion parameter multi-modal modal named Pixtral \cite{agrawal2024pixtral12b} that maintains strong NLP performance while incorporating vision. Pixtral has been shown to outperform larger models on many tasks across modalities such as MM-MT-Bench.
\section{Trends in Large Language Models for EDA}

This section presents a meta-study of how LLMs have been employed in CAD in the recent years.
It reviews all publications in the selected venues, focusing on studies where LLMs are utilized for CAD, as outlined in the introduction.
Our criterion excludes a large number of works that use LLMs as an application (accelerators, approximate computing, etc.).
Only regular papers are considered while invited papers are excluded.
We study the years 2022 to 2024.
This meta-study answers the following main questions:

\begin{itemize}
\item Which areas in CAD are well-explored / unexplored with LLMs?
\item Which techniques have been used and proven effective?
\item Which are the observable trends?
\end{itemize}

In this survey, we concentrate on the pivotal stages of EDA that span from system-level design to the physical implementation of the chip.
Specifically, our exploration is anchored in the following key design phases.

1) \textbf{System-level design}:
This initial stage involves transforming a high-level specification of the IC into an RTL description.
System-level design optimizes non-functional properties, guiding decisions on hardware versus software implementations, processor configurations, and the scheduling and binding of operations.
This phase is critical in defining the system's architecture and operational blueprint.

2) \textbf{RTL design}: RTL description of a circuit is created using HDLs such as Verilog or VHDL, which represent the data flow between registers and the logical operations performed.
This phase defines the functional behavior of the circuit and serves as input for synthesis.

3) \textbf{Synthesis and Physical Design}: Logic synthesis transforms the RTL description of a circuit to a gate-level representation in the target technology.
Physical design includes placing the logic gates on the die, routing the connecting nets, design of the clock trees, and building a power/ground network. The output of the physical design phase is a geometric representation of the circuit.

4) \textbf{Analog circuit design}: Analog design handles continuous signals for applications like amplifiers, filters, and converters. This process involves selecting topologies and device sizing to meet specifications for gain, bandwidth, and noise, often requiring precise layout design for mixed-signal environments.




\subsection{System-level Design}

 \cite{yan2023viability} explores the use of Large Language Models (LLMs) for software-hardware (SW-HW) co-design, focusing on the design of Compute-in-Memory (CiM) DNN accelerators.
It introduces the LCDA framework, the first of its kind to utilize LLMs for the co-design of accelerators for deep neural networks.
The study demonstrates the efficacy of LCDA by co-designing DNN topology and CiM DNN accelerators, achieving a substantial 25x speedup compared to state-of-the-art methods while maintaining similar performance levels.
This speedup is due in large part to the ability of the pretrained LLM to intelligently decide on design attributes avoiding the "cold-start" problem traditional co-design optimizers face when random guessing design specifications.
The work also highlights the potential of leveraging LLMs and the LCDA framework for SW-HW co-design, pointing out future research directions such as explainable NAS and the development of open-source LLMs tailored for co-design purposes.
The proposed approach addresses the challenges of deploying DNNs on edge devices by accelerating the design process, ultimately enabling the rapid and efficient deployment of DNNs on edge platforms with limited computational resources and power constraints.

\cite{li2024specllm} focuses on generating and reviewing VLSI design specifications using Large Language Models (LLMs), addressing the initial and fundamental stage of the integrated circuit (IC) design process.
The paper categorizes architecture specifications into three distinct abstraction levels and explores LLM applications in both writing specifications from scratch and converting RTL code into detailed specifications, as well as reviewing existing architecture specifications.
This approach aims to enhance efficiency and accuracy in the crucial aspect of chip design, making it pertinent to discussions on system-level circuit design.

GPT4AIGChip\cite{fu2023gpt4aigchip} proposed a framework aimed at democratizing AI accelerator design by using natural language instructions instead of specialized hardware languages, automating demo-augmented prompt-generation pipeline to guide LLMs in producing high-quality AI accelerator designs.

Chip-Chat\cite{blocklove2023chip} used ChatGPT-4 to perform all the choices in the design of a microprocessor, from the initial design specification to tapeout.
The LLM acting in a conversational environment with an experienced hardware designer who was only acting to validate its choices was able to produce viable and quality design specifications at both the high system level and the component level.
This work acts as an example of how LLMs can be used in all parts of the design cycle including system-level specification with extremely minimal initial input.

A comparative analysis of these LLM-driven approaches reveals distinct trade-offs in their application to system-level design.
LCDA\cite{yan2023viability} demonstrates superior optimization efficiency with its 25x speedup, but requires careful curation of the design space and constraints to maintain performance.
Chip-Chat\cite{blocklove2023chip} showcases end-to-end design capabilities with minimal human intervention, achieving RTL-level specifications, though its reliance on GPT-4 may limit reproducibility and cost-effectiveness for large-scale deployment.
SpecLLM\cite{li2024specllm} offers more focused specification generation and review capabilities across abstraction levels, potentially providing better quality control but with a narrower scope compared to end-to-end solutions like Chip-Chat.
GPT4AIGChip\cite{fu2023gpt4aigchip}'s demo-augmented approach strikes a balance between automation and quality control, though its effectiveness heavily depends on the quality and coverage of the demonstration examples.
Notably, while LCDA\cite{yan2023viability} and GPT4AIGChip\cite{fu2023gpt4aigchip} provide quantitative performance metrics, Chip-Chat\cite{blocklove2023chip} and SpecLLM\cite{li2024specllm}'s evaluations are more qualitative, focusing on design viability and specification quality.
This suggests a need for standardized evaluation metrics across different LLM-based design approaches.

These papers present groundbreaking approaches to leveraging LLMs for system-level design challenges, highlighting both the current potential and future directions in this field.
The application of LLMs in software-hardware co-design, particularly in the development of CiM DNN accelerators, showcases a promising path toward significantly speeding up the design process while maintaining high performance, a critical advancement for deploying DNNs on edge devices.
On the other hand, using LLMs for generating and reviewing VLSI design specifications at various abstraction levels demonstrates a novel approach to enhancing the efficiency and accuracy of the initial stages of IC design.
These studies underscore the promising role of LLMs in automating and optimizing the design of system level specifications, yet also point to challenges such as the need for explainable models and domain-specific LLM development.
Future research could focus on improving the interoperability of these models, expanding their application areas, and creating open-source, domain-tailored LLMs to foster broader adoption and innovation.

In the realm of system-level design utilizing LLMs, challenges abound, including the imperative for explainable models to enhance transparency and trust in automated design decisions. The creation of domain-specific LLMs is critical to address unique design intricacies, while maintaining a balance between speed and performance in DNN accelerator design remains a pivotal concern. Additionally, while LLMs can greatly accelerate the design specification process, ensuring the accuracy and efficiency of LLM-generated design specifications is paramount.

Promising directions for overcoming these challenges include the exploration of explainable Neural Architecture Search (NAS), which aims to demystify model decisions. The development of open-source, domain-adapted LLMs promises to democratize access and spur innovation in system-level design. Leveraging LLMs for initial system design and rapid prototyping software-hardware co-design could significantly shorten development cycles. Furthermore, enhancing LLM capabilities for generating, converting, and reviewing detailed design specifications across abstraction levels could revolutionize the initial stages of IC design.

Another significant stride could be made by accurately capturing the variations introduced during the backend implementation phase. This approach requires foundational LLMs tailored to specific computing architectures of design companies and tech node-specific models provided by foundries. The tech node foundational LLM model would encapsulate the Power, Performance, and Area (PPA) characteristics for a broad spectrum of designs, sizes, and types. Concurrently, the system-level model, offered by the design company, would estimate performance based on given workloads and system specifications. Integrating these models has the potential to profoundly revolutionize the initial stages of IC design, significantly reducing the time and resources needed for Engineering Change Orders (ECO) cycles. This integrated approach promises to streamline the design process, enhancing efficiency, reducing costs, and accelerating innovation in IC design.

\subsection{RTL Design}


RTL design has emerged as a prominent focus area for LLM applications in EDA, encompassing multiple aspects including \textbf{RTL code generation, debugging, correction, and verification (e.g., testbench generation)}.
Recent research in applying LLMs to RTL design has coalesced around three main trends.
First, researchers are exploring advanced prompt engineering techniques and developing automated flows that take advantage of existing commercial LLMs.
Second, researchers are investigating domain adaptation approaches that go beyond text input modifications, incorporating supervised fine-tuning, data engineering, and tokenizer customization methods.
Third, researchers have proposed LLM autonomous agent methods that iteratively improve code generation by incorporating feedback from user-defined tools.

\textbf{Prompt engineering} for EDA involves augmenting effective inputs to guide the LLMs.
One challenge that is commonly identified by the papers that involves prompt engineering for EDA tasks is that it is hard for current LLMs to understand lengthy inputs which exhibit long dependencies.
In particular, for closed-source commercial LLMs, the scarcity of domain-specific languages such as HDLs or TCL scripts in their pretraining data makes the challenge even harder.
There have been various efforts to address this challenge.
\cite{nair2023generating} studies a diversity of prompts for using ChatGPT to generate hardware code that is secure and resistant to common vulnerabilities listed in the Common Vulnerability Enumerations (CWEs), particularly focusing on the hardware design perspective.
\cite{dehaerne2023deep} proposes a novel deep learning framework aimed at training models for Verilog code auto-completion, leveraging a specially curated Verilog dataset to significantly enhance the efficiency and quality of digital circuit design and verification processes. 
\cite{du2023power} explores the utilization of LLMs for enhancing the FPGA-based development of complex signal-processing algorithms in wireless communication systems, demonstrating their potential in code generation, refactoring, and validation, particularly through the successful generation of a 64-point Verilog FFT module using advanced prompting techniques.
\cite{sandal2024zero} introduces a novel approach to zero-shot RTL code generation using LLMs enhanced with an attention sink mechanism, demonstrating the ability to generate functional, optimized, and industry-standard compliant RTL code from high-level design specifications, thereby significantly advancing the automation of hardware design processes.
\cite{tsai2024rtlfixer} utilizes RAG and ReAct prompting techniques to autonomously debug and correct syntax errors in Verilog code.
\cite{chang2023chipgpt} proposed ChipGPT, a framework that uses natural language specifications for automatic chip logic design, integrating language models into EDA tools without retraining.
The authors present a four-stage zero-code logic design framework based on LLMs, demonstrating an automated design environment that generates hardware logic designs from natural language specifications.
PyHDL-Eval\cite{batten2024pyhdl} is the first LLM evaluation framework specifically designed for hardware design using Python-embedded domain-specific languages (DSLs).
\cite{batten2024pyhdl} finds that LLMs generally perform better with Verilog despite Python's wider presence in training data, while also showing the critical importance of in-context learning for improving performance with Python-embedded DSLs.
VerilogReader\cite{ma2024verilogreader} integrates LLMs into Coverage Directed Test Generation by introducing two key components: a Coverage Explainer that reformats coverage data into LLM-readable format and a DUT Explainer that provides design context through natural language descriptions.
The framework demonstrates superior performance over random testing for simple/medium designs while revealing current LLM limitations on complex hardware designs (>64 states), suggesting the need to combine LLMs with structural approaches like GNNs for industrial applications.

\textbf{Domain-adaptive pre-training (DAPT)} and \textbf{supervised fine-tuning (SFT)} is another dominant approach which customizes the model by alternating its parameters for better domain adaptation.
\cite{liu2023rtlcoder} introduces RTLCoder, an open-source, efficient LLM tailored for automatic RTL code generation, achieving superior performance to GPT-3.5 with a modest parameter count and leveraging a unique dataset and training scheme for improved accuracy and efficiency in digital circuit design tasks.
VerilogEval\cite{liu2023verilogeval} is another comprehensive benchmarking framework and dataset for evaluating the performance of LLMs in generating Verilog code for hardware design and verification, showcasing the potential to enhance LLM Verilog coding capabilities through supervised fine-tuning with synthetic problem-code pairs.
An alternative approach to domain adaptation without touching the parameters of an LLM is to customize its decoding process.
\cite{delorenzo2024make} presents an automated technique for generating high-quality, optimized Verilog RTL code by integrating Monte Carlo Tree Search (MCTS) into the token generation process, significantly improving functional correctness and PPA efficiency.
\cite{goh2024english} uses supervised fine-tuning of the Mistral-7B model on Verilog datasets, coupled with dataset reshuffling to expand and diversify training examples, resulting in significant accuracy improvements in Verilog code generation.
MEIC\cite{xu2024meic} proposed to automate RTL debugging with a dual-agent architecture where debugging and scoring were respectively assigned to a GPT-4 Turbo agent.
The framework applied self-planning and role-based prompt engineering to break down complex debugging tasks while fine-tuning the model with domain knowledge through system-level instructions about Verilog standards.
It focused on iterative improvement through its scoring mechanism and rollback capabilities to handle LLM hallucinations.

\begin{table*}[tb]
\centering
\caption{HDL generation researches focusing on prompt engineering techniques.}
\begin{tabular}{>{\centering\arraybackslash}m{2cm}|>{\centering\arraybackslash}m{2cm}|>{\centering\arraybackslash}m{2cm}|m{9cm}}
\hline
Project Name & Model & Method & \multicolumn{1}{c}{Description} \\
\hline
\hline
ChipGPT\cite{chang2023chipgpt} & ChatGPT (GPT-3.5) & In-context Learning & ChipGPT proposes a scalable, zero-code logic design framework using in-context learning with GPT-3.5, without retraining or fine-tuning. It enhances the generation of Verilog code from natural language specifications, focusing on improving programmability, controllability, and design optimization. The framework includes a prompt manager and output manager to optimize the design process, demonstrating advancements in natural language-based hardware design. \\
\hline
An Empirical Evaluation\cite{schafer2023empirical} & GPT-3.5-turbo & Prompt Crafting & This study evaluates the effectiveness of LLMs, specifically GPT-3.5-turbo, in generating automated unit tests without additional training. By crafting detailed prompts, including function signatures and documentation, the approach achieves significant coverage in JavaScript API testing, demonstrating the LLM's capability to produce meaningful and diverse test cases with minimal manual intervention. \\
\hline
RTLLM\cite{lu2023rtllm} & GPT-3.5 with Self-Planning & Benchmarking and Prompt Engineering & RTLLM introduces an open-source benchmark for RTL generation, focusing on syntax correctness, functionality, and design quality. It demonstrates the efficacy of a novel prompt engineering technique, self-planning, significantly enhancing GPT-3.5's performance in generating design RTL from natural language instructions, highlighting improvements in handling complex hardware design tasks. \\
\hline
AutoChip\cite{thakur2023autochip} & Multiple LLMs including GPT-3.5-turbo & Iterative Feedback & AutoChip automates HDL generation by iteratively refining Verilog code using feedback from compilation and simulation errors. It leverages various LLMs to improve code accuracy through multiple rounds of feedback, demonstrating a novel approach to reducing human intervention in Verilog module generation. \\
\hline
Chip-Chat\cite{blocklove2023chip} & ChatGPT-4 & Conversational Co-Design & "Chip-Chat" explores conversational co-design with ChatGPT-4 for hardware development, achieving a novel 8-bit accumulator-based microprocessor design. This work highlights the potential of conversational LLMs in hardware design, providing insights into leveraging AI for innovative design approaches and the practical application of LLMs in co-architecting complex hardware, culminating in a successful tapeout. \\
\hline
VeriGen\cite{thakur2023verigen} & CodeGen-16B & Fine-tuning on Verilog corpus & VeriGen explores automating Verilog code generation by fine-tuning the CodeGen-16B model on a large Verilog corpus from GitHub and textbooks. It demonstrates improved syntactic correctness and functional performance of generated Verilog code, highlighting the potential of fine-tuned LLMs in hardware design automation. The approach achieves competitive performance against commercial models like GPT-3.5-turbo, showcasing the effectiveness of domain-specific fine-tuning. \\
\hline

\hline

\end{tabular}
\label{tab:hdl-gen-prompt}
\end{table*}


Table~\ref{tab:hdl-gen-prompt} summarizes recent research efforts in HDL generation using LLMs, showcasing a broad spectrum of methodologies from fine-tuning with EDA tools to domain-specific adaptations and conversational co-design.
The table reflects a significant trend towards leveraging LLMs to enhance efficiency and innovation in hardware design.
Projects like AutoChip and ChipNeMo exemplify the promising directions of utilizing feedback from EDA tools and domain-specific customizations, including custom tokenizers and retrieval models, respectively.
These approaches highlight the potential for LLMs to significantly reduce manual intervention in the design process, even though complete automation remains challenging, particularly in verification stages.
Moreover, the table reveals untapped opportunities in multi-modality on graph-based RTL abstraction and the novel idea of re-encoding RTL code into formats closer to natural language to bridge the gap between human designers and machine processes.
Despite the advancements, challenges such as scalability with modern complex designs and the irreplaceable role of human oversight in verification emphasize the need for continued innovation and research in this field.
The inclusion of insights on promising research directions and challenges enriches the understanding of the current landscape and future potential in HDL generation using LLMs, highlighting both the advancements made and the hurdles that lie ahead.

The integration of LLMs in hardware design automation, as outlined in Table~\ref{tab:hdl-gen-prompt}, highlights promising research directions and inherent challenges.
Utilizing feedback from EDA tools and domain-specific customizations, such as seen in AutoChip and ChipNeMo, showcases the innovative use of LLMs to refine hardware designs iteratively and generate domain-specific HDL with higher accuracy.
These approaches indicate a significant step towards minimizing manual intervention by adapting LLMs to the intricacies of hardware design language and processes.
The exploration of multi-modality on graph-based RTL abstraction and the re-encoding of RTL code into formats closer to natural language suggests future paths for making hardware design more intuitive and aligned with LLM capabilities, potentially enhancing the design process's efficiency and creativity.

However, the challenges of ensuring scalability in modern complex designs and the necessity of human intervention, especially in verification stages, remain significant hurdles.
While LLMs can reduce manual effort by generating more accurate and context-aware HDL, the nuanced understanding and decision-making capabilities of human designers are irreplaceable.
These challenges underscore the need for ongoing innovation in LLM research focused on hardware design, aiming to balance the strengths of automated tools with the critical oversight provided by human expertise.
As the field evolves, continued advancements in LLM capabilities and methodologies hold the promise of further transforming hardware design automation, making it more efficient and accessible while navigating the complexities of modern technological demands.


\begin{figure}[bt]
    \centering
    \includegraphics[width=1.1\textwidth]{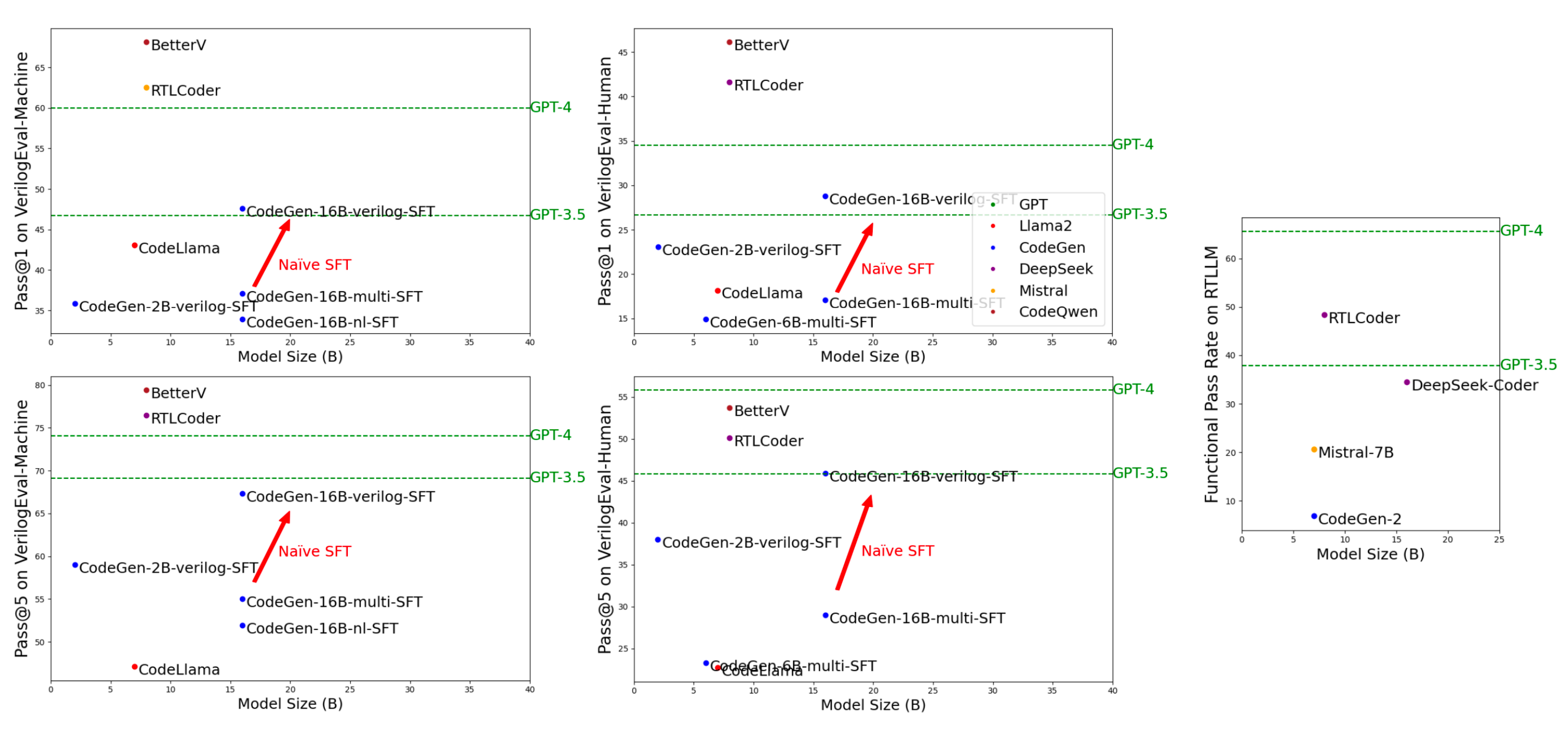}
    \caption{Comparing the functional correctness of Verilog code generation on VerilogEval-machine~\cite{liu2023verilogeval}/ VerilogEval-Human~\cite{liu2023verilogeval}/ RTLLM~\cite{lu2023rtllm}. The performance of the naive SFT methods correlates well with the model size. State-of-the-art customization methods have outperformed the best of prompt engineering methods with GPT-4.}
    \label{fig:hdl_gen_comparison}
\end{figure}

Figure~\ref{fig:hdl_gen_comparison} compares the functional correctness of generated Verilog codes from multiple research works\cite{liu2023rtlcoder,pei2024betterv,liu2023verilogeval,liu2023chipnemo,nijkamp2022codegen}.
Current state-of-the-art models~\cite{pei2024betterv,liu2023rtlcoder} require more sophisticated methods than supervised fine-tuning alone.
Specifically, BetterV\cite{pei2024betterv} decomposed the conditional probability of the token during generation using Baysian rule and used a generative discriminator to modify the conditional probability, thus controlling the token generation to prefer desired tokens.
RTLCoder~\cite{liu2023rtlcoder} adopted a different, data-oriented approach of assigning scores to the RTL design candidates, which are further fed to the model during the fine-tuning process.
Both BetterV and RTLCoder developed special techniques that differentiate Verilog examples with different qualities.
These techniques are key to the achieved improvement compared to other works~\cite{liu2023verilogeval} that adopted simpler SFT methods.
Interestingly, with naive SFT approaches, the performance of these models in terms of functional correctness tends to increase as the model size grows.
Figure~\ref{fig:hdl_gen_comparison} also shows the importance of the pre-training corpus in the Verilog generation problem.
Comparing the performance between \texttt{CodeGen-16B-nl-SFT}, \texttt{CodeGen-16B-multi-SFT}, and \texttt{CodeGen-16B-verilog-SFT}, we noticed that pre-training on a multi-programming language (\texttt{multi}) corpus shows minimal advantage over training on a natural language (\texttt{nl}) corpus.
And pre-training on a small Verilog corpus significantly boosts the performance, despite that all models performed supervised fine-tuning on Verilog data.
Another observation is that using prompt engineering with closed-source LLMs (e.g., ChatGPT) remains highly competitive against other fine-tuned LLMs.

\begin{table*}[tb]
\centering
\caption{HDL generation researches focusing on training or fine-tuning techniques.}
\centering
\begin{tabular}{>{\centering\arraybackslash}m{2cm}|>{\centering\arraybackslash}m{2cm}|>{\centering\arraybackslash}m{2cm}|m{9cm}}
\hline
\hline
Project Name & Foundation Model & Method & \multicolumn{1}{c}{Summary} \\
\hline
\hline
VeriGen\cite{thakur2023verigen} & CodeGen-16B & SFT on Verilog & It fine-tunes CodeGen-16B on a substantial Verilog corpus from GitHub and textbooks. Its performance matches GPT-3.5-turbo. \\
\hline
ChatEDA\cite{he2023chateda} & Llama-20B & Instruciton Tuning with QLoRA & It collects EDA instructions by applying self-instruction paradigm on GPT-4, and uses the collected instructions to fine-tune a Llama-20B model.\\
\hline
ChipNemo\cite{liu2023chipnemo} & Llama-7B/13B/70B & SFT with domain-specific instructions & Its domain adaptation techniques cover from tokenization to SFT, and demonstrated the adapability results on various EDA tasks.\\
\hline
RTLCoder\cite{liu2023rtlcoder} & Mistral-7B & SFT with Verilog scoring & It proposed an automated GPT-based Verilog dataset generation flow, and further fine-tuned an LLM on scored Verilog examples.\\
\hline
\cite{dehaerne2023deep} & CodeGen-345M & SFT on Verilog & It fine-tunes CodeGen-345M on publicly available Verilog data.\\
\hline
BetterV\cite{pei2024betterv} & CodeLlama-7B-Instruct + TinyLlama & SFT and controlled generation  & It uses a TinyLlama-based Verilog discriminator to guide the CodeLlama-based Verilog generator model. It has achived state-of-the-art in the VerilogEval-machine\cite{liu2023verilogeval} benchmark.\\
\hline
\hline

\end{tabular}
\label{tab:hdl-gen-sft}
\end{table*}

\textbf{LLM autonomous agent framework} is another promising way to improve the code generation of RTL design.
RTLFixer~\cite{tsai2024rtlfixer} develops a debugging framework using Retrieval-Augmented Generation (RAG) and ReAct~\cite{yao2023react} prompting. In detail, they create a database that categorizes different compile error types and annotate human expert guidance with each type. Also, ReAct prompting, which consists of thought, action, and observation components, can indicate LLM to generate instruction and observation based on compiler's feedback. Thus, LLM can read the compiler's result and query the database to effectively debug its previous generated RTL code.
Then, VerilogCoder~\cite{ho2024verilogcoder} develops a novel task-based planning technique to extract circuit signal and transition and an abstract syntax tree (AST)-based waveform tracing tool to let LLM backtrace signals to find functional mistakes. 
Notably, based on VerilogEval-Human v2~\cite{pinckney2024revisiting}, VerilogCoder can achieve pass rate of 94.2\% (67.3\%) using GPT4-turbo (Llama3), while using GPT4-turbo (Llama3) without its agent framework only achieve 60.3\% (41.7\%) pass rate.
This result also suggests that the close-sourced model e.g., GPT4-turbo is more powerful than open-sourced model Llama3 in the agentic framework.
VerilogCoder also largely outperforms state-of-the-art open-sourced models RTLCoder~\cite{liu2023rtlcoder}, DeepSeek coder~\cite{guo2024deepseek}, and CodeGemma~\cite{google2024codegemma}.

\subsection{Logic Synthesis and Physical Design}

Logic synthesis and physical design serve as the cornerstone of circuit generation and have garnered significant attention in the development of large language model (LLM)-assisted applications in this domain. Two key trends have emerged in applying LLM methods to the logic design stage: one focuses on leveraging LLMs for Tcl script generation, while the other emphasizes enhancing documentation question-and-answer (Q\&A) systems.

\textbf{Script Generation} has become an important area where LLMs are utilized to streamline electronic design automation (EDA) workflows. 
ChatEDA\cite{he2023chateda}, leverages large language models (LLMs) like GPT-4 and AutoMage to streamline the electronic design automation (EDA) process. This system provides a conversational interface that improves productivity and usability by enabling designers to interact directly with the design workflow through task planning, script generation, and execution. The authors highlight the significance of interfacing EDA tools using LLM to unleash circuit design productivity, and they propose ChatEDA as a solution to inspire next-generation EDA tool evolution. ChipNemo\cite{liu2023chipnemo} also introduces LLMs for script generation with industrial chip design flow. Besides purely using LLM as a chatbot and generating the EDA script for synthesis and physical implementation, this work also presents its potential application in bug summarization and analysis. Evaluated with different LLM with 7B/ 13B/ 70B, and with various domain adaptation techniques such as custom tokenizers, domain-adaptive pretraining, supervised fine-tuning, and domain-adapted retrieval models, the study highlights the effectiveness of these domain-specific adaptations, achieving performance comparable or superior to much larger models.

\textbf{Documentation Q\&A}
is another rising application beyond script generation to enable chip designers to efficiently retrieve accurate and relevant information from the extensive and complex documentation associated with EDA tools. retrieval-augmented generation (RAG) framework is a common technique to enhance LLM Q\&A ability regarding documentation Q\&A. While existing RAG systems struggle with EDA-specific terminologies and complex workflows, several customized RAG methods can be proposed to enhance LLM understanding under the EDA corpus and enhance their ability to generate accurate answers. 
To support these advancements, several datasets have been introduced to provide a foundation for LLM training and evaluation in the EDA domain. For example, EDA Corpus~\cite{EDA_Corpus} and ORD-QA~\cite{pu2024customizedretrievalaugmentedgeneration} include question-answer pairs and prompt-script pairs tailored for EDA tasks. In addition to dataset development, customized RAG techniques have been proposed specifically for the EDA domain. Openroad-assistant~\cite{Openroad_assistant} leverages a retrieval-aware fine-tuning technique to enhance their LLM framework for question-answering. RAG-EDA~\cite{pu2024customizedretrievalaugmentedgeneration} introduces a customized retriever and enhanced reranker to enhance LLM understanding regarding EDA terminologies and workflows.

Looking ahead, several promising directions could further extend the utility of LLMs in VLSI physical design.
The generation and evaluation of netlists represent a crucial area where LLMs could provide significant benefits, automating a task that is both foundational and complex.
Additionally, supporting layout features through multi-modality—integrating text with graphical data—could enable LLMs to better understand and contribute to the layout design process.
Encoding backend libraries for design Power, Performance, and Area (PPA) prediction could revolutionize how designers approach optimization, allowing for more efficient and accurate design choices.
Moreover, the use of Retrieval-Augmented Generation (RAG) for automated design analysis and debugging could streamline the identification and resolution of design issues, significantly accelerating the development cycle.

However, these potential advancements are not without their challenges, with domain adaptation for the backend design process standing as a primary hurdle.
The backend phase of VLSI design, encompassing tasks such as placement, routing, and optimization, involves a complex interplay of design rules, performance metrics, and physical constraints.
Adapting LLMs to effectively understand and generate solutions within this domain requires not only advanced training techniques but also deep integration of domain-specific knowledge.
A particular challenging application is timing estimation, where LLMs must capture intricate relationships between physical layout and parasitic effects while maintaining the mathematical precision of traditional static timing analysis tools.
Looking forward, LLMs could potentially accelerate timing convergence by learning patterns from historical design iterations to suggest targeted optimizations while leveraging their pattern recognition capabilities to proactively identify and prioritize critical timing paths that are likely to cause violations based on similar design contexts learned by the models.
These challenges underscore the need for continued research and collaboration between the fields of artificial intelligence and VLSI design, aiming to develop models that are not only technically proficient but also deeply attuned to the unique requirements of backend design processes.
It is noteworthy that FPGA design and deployment optimization also faces unique challenges distinct from ASIC design, particularly in resource allocation, routing complexity, and design space exploration, where the reconfigurable nature of FPGAs creates both opportunities and constraints.
While research on LLM applications in this domain remains limited, promising directions include leveraging LLMs for high-level design planning, automated constraint generation, intelligent design space exploration, and tool chain optimization, potentially streamlining the FPGA development process through pattern recognition and learning capabilities.



\subsection{Analog Circuit Applications}

Analog circuits are an intricate process, traditionally involving multiple steps from initial specification to final verification. Each step requires specialized technical experience and iterative refinement to meet stringent performance criteria. Analog design automation, typically using ML~\cite{mina2022review}, is a promising direction to accelerate or even automate circuit development without human intervention.
The advent of LLMs offers promising approaches to enhance the efficiency, creativity, and outcome of each stage of the analog circuit design process. In the following, we discus the potential of LLM and some recent works to revolutionize each step of analog design automation.


\noindent \textbf{Initial Specification and Topology Selection} 
The design process begins with defining precise specifications for the circuit, such as gain, bandwidth, and power consumption, followed by the selection of a circuit topology.
Currently, this step requires human experience to process manually on the basis of previous experience. 
LLMs can contribute to this initial phase by analyzing extensive databases of design precedents.
This allows them to suggest suitable specifications and recommend circuit topologies that possibly aligned with the desired performance.
DocEDA~\cite{chen2024docedaautomatedextractiondesign} uses LLM and computer vision techniques to automatically extract and analyze layouts, parameters, and circuit topologies from the circuit technical documents. Then, they use the extracted database for RAG to optimize topology selection and user queries related to analog circuits.

\noindent \textbf{Schematic Design and Simulation}
With specifications and topology in hand, designers move on to creating detailed schematic designs. 
By leveraging LLM's training on diverse designs, LLMs can automate the generation of initial schematic designs. 
Additionally, using in-context learning, LLM can propose circuit optimizations, thereby reducing manual effort and expertise dependency.  
Furthermore, in the simulation and analysis phase, LLMs can help analyze error reports from simulation tools and guide designers toward solutions that meet specifications more efficiently.
AnalogCoder~\cite{lai2024analogcoder} designs a feedback-enhanced method for topology generation, which LLM reads the functional check and simulation results to iteratively refine the generated circuits and enlarge the circuit library for retrieval for future generation.
ADO-LLM~\cite{yin2024ado} incorporates LLM with Bayesian optimization to infuse the domain knowledge given by LLM in the search process to enhance effectiveness.
LaMAGIC~\cite{pmlr-v235-chang24c} performs supervised finetuning and explore several circuit formulations for topology generation to let LLM learn for relations between circuits and simulation results to achieve high success rates in a one-shot generation manner.

\noindent \textbf{Layout Design, Post-Layout Simulation}
The layout design phase involves placing and routing components on the silicon chip to minimize unwanted parasitic effects and enhance circuit performance and need the subsequent post-layout simulation, which accounts for physical phenomena not considered in schematic simulations, to validate the solution. 
Compared to VLSI physical design, analog layout design must satisfy more complicated constraints, such as the symmetry requirement and the wire matching between specific components. 
These constraints are often designed by human based on their expertise. 
Thus, LLMs can offer recommendations on layout constraints to help designer selecting promising constraints that towards successful layouts.
LLANA~\cite{chen2024llm} combines LLM with Bayesian optimization by using the few-shot generation ability of LLM to enhance the generation of layout constraints.
LayoutCopilot~\cite{liu2024layoutcopilot} develops a multi-agent LLM system by combining the high-level design insights of LLM with human interactive feedback to improve the analog layout performance.

While LLMs hold great promise for revolutionizing analog circuit design automation, a primary challenge remains: training these models on extensive and diverse datasets of analog designs. The effectiveness of an LLM hinges on the quality and breadth of the data it has learned from. 
Given the specialized and sometimes proprietary nature of analog circuit designs, compiling a comprehensive dataset that covers a vast range of design strategies, topologies, and technologies is a challenging task. 
Moreover, the depth of understanding required to navigate the complexities of analog design complicates the training process, necessitating datasets that are not only large but also richly detailed with design logic, performance metrics, and optimization strategies.

\section{From the LLM Methodology Perspective}


This section delves into the methodology behind the application of LLMs for EDA, focusing on model architecture and size, customization techniques, and feature representation. It also provides an overview of the current academic infrastructure supporting LLM research in EDA, including open-sourced models and benchmarks/datasets.

\subsection{Selection of LLM and Size}



The choice of different LLMs and their corresponding model size are critical factors in their ability to process and generate complex, domain-specific information pertinent to EDA tasks.
Larger models, such as GPT-3.5 and its successors, have demonstrated improved performance in understanding intricate design details and generating high-quality HDL code~\cite{liu2023rtlcoder}.
The choice of LLM architectures primarily lies in LLaMA\cite{touvron2023llama}/LLaMA-2\cite{touvron2023llama2}, CodeGen\cite{nijkamp2022codegen}, Mistral\cite{jiang2023mistral}, and GPT-3/3.5/4\cite{brown2020language}.

Different pretrained LLMs employ various transformer architectures. The core of transformer architectures includes encoder, decoder modules.
Encoder modules enable the full visible attention mechanism, so they are adept at processing input data and generating a comprehensive, context-rich representation.
This characteristic makes encoder models particularly suitable for tasks requiring a deep understanding of complex input data, such as parsing and analyzing HDL code or optimizing chip layouts based on design specifications.
On the contrary, decoder modules apply a causal attention mask from left to right to enable models to process tokens on the past. This makes it excel at generating output based on the encoded representations.

Different combinations of encoder and decode modules form the encoder-only, decoder-only or encoder-decoder~\cite{raffel2020exploring} transformer architectures.
Encoder-only model~\cite{devlin2018bert} is primarily used for language prediction task, e.g., sentiment prediction. 
Decoder-only (e.g., GPT and LLaMA series) or encoder-decoder (e.g., T5~\cite{raffel2020exploring}) models are mainly deal with real-world generation tasks. 
Specifically, encoder-only models are shown to be more effective compared to encoder-decoder ones in recent research advancement.
Additionally, there is a new architecture called PrefixLM~\cite{raffel2020exploring} that enables full visible attention in prior input tokens and masks the attention of future tokens in generation. 
PrefixLM is good at tasks that need to highly understand the full user input, so CodeGen series are training this architecture for code generation.


Table~\ref{tab:hdl-gen-prompt} and~\ref{tab:hdl-gen-sft} provides a concrete illustration of how different projects leverage these LLM aspects to address EDA challenges.
The works exemplify the strategic deployment of different pretrained LLMs and  model sizes to optimize the generation of hardware description languages, each tailored to specific facets of the EDA workflow.
The studies in Table~\ref{tab:hdl-gen-prompt} explore various pretrained LLMs employing in-context learning for EDA tasks, revealing diverse performance levels across tasks. This variation underscores the importance of selecting appropriate pretrained LLMs for specific EDA applications. However, each study only explores a limited number of LLM options. There is a noticeable gap in comprehensive research on selecting the optimal pretrained LLM for EDA tasks.

Moreover, as shown in Table~\ref{tab:hdl-gen-sft}, some projects have fine-tuned LLMs with domain-specific data, leading to significant performance improvements in their respective tasks. This suggests that fine-tuning LLMs with EDA-specific knowledge is a promising avenue for enhancing LLM capabilities in this field.
It is also observed that few studies have explored scaling model sizes to 70 billion parameters and beyond. The likely reason is that increasing model size must coincide with expanding the dataset size to see corresponding improvements in model performance. The existing VerilogEval dataset, containing only 8,000 samples, is relatively small.
Since enlarging model sizes is a common strategy in other LLM applications, pursuing research to expand the VerilogEval dataset and increase LLM model sizes presents a promising direction to advance LLM applications in EDA.

Finally, when developing LLM autonomous agent framework~\cite{tsai2024rtlfixer, ho2024verilogcoder, chang2025drccoder, ho2024large} for hardware tasks, the close-sourced models, e.g., GPT4, GPT4-turbo, GPT4o, are better choices compared to other open-sourced models, e.g., Llama3. 
The close-sourced models have great abilities to interact with user-defined tool functions and perform iterative improvement based on their previous generation rounds.

\subsection{Customization Techniques}

\begin{table}[tb]
\centering
\caption{Different datasets for fine-tuning towards LLM4EDA.}
\begin{tabular}{|l|l|l|l|}
\hline
Dataset & Task & Size & Description \\
\hline
ChipNeMo\cite{liu2023chipnemo} & Domain Adaptation & 24.1 B tokens & Data from NVBugs and other sources for EDA tasks. \\
ChatEDA\cite{he2023chateda} & Instruction Tuning & 1,500 instructions & Instructions for assistant chatbots in EDA. \\
GPT4AIGChip\cite{fu2023gpt4aigchip} & Code Generation & 7,000 snippets & Open-source HLS code snippets for HDL generation. \\
VeriGen\cite{thakur2023verigen} & Educational Material & 400 MB & From Verilog textbooks for verification and analysis. \\
VerilogEval\cite{liu2023verilogeval} & Evaluation Set & 8,502 samples & Designs generated for verification and evaluation. \\
RTLCoder\cite{liu2023rtlcoder} & Design Generation & 10,000 designs & Generated from GPT-3.5 for RTL coding and synthesis. \\
\hline
\end{tabular}
\label{table:dataset}
\end{table}

Diving into the customization techniques used by various projects highlights the critical role of adapting LLMs to the nuanced domain of EDA tasks. Techniques such as fine-tuning, domain-adaptive pretraining, supervised fine-tuning, and innovative prompt engineering are essential to leverage LLMs effectively in hardware design and verification.
We summarize the techniques here:
\begin{itemize}
    \item \textbf{Fine-tuning with EDA tools}, as seen in ChatEDA and VeriGen, enhances models' understanding and generation capabilities for specific EDA tasks like script generation and HDL code production.
    \item \textbf{Domain-adaptive pretraining and supervised fine-tuning}, exemplified by ChipNeMo, tailor models such as LLaMA2 for chip design, employing custom tokenizers and retrieval models for improved performance.
    \item \textbf{Innovative prompt engineering techniques},  used in RTLLM, ChipGPT, and GPT4AIGChip, demonstrate the versatility of in-context learning and prompt design in enhancing LLM output for complex hardware design tasks.
    \item \textbf{Autonomous agent framework}: Established in RTLFixer, VerilogCoder, and DRC-Coder~\cite{chang2025drccoder}, this framework demonstrates the capability of LLMs to iteratively enhance hardware code generation. It achieves this by interacting with customized tool functions, eliminating the need for finetuning and effectively addressing complex hardware challenges.
    \item \textbf{Retrieval augmented generation}: Exemplified by Openroad-assistant\cite{Openroad_assistant} and RAG-EDA\cite{pu2024customizedretrievalaugmentedgeneration}, this technique enhances LLM's capability to handle complex EDA documentation Q\&A tasks. Through specialized retrieval and reranking mechanisms optimized for EDA terminology, coupled with retrieval-aware fine-tuning, these systems demonstrate superior performance in accessing and interpreting technical documentation compared to traditional RAG frameworks.
\end{itemize}
These strategies illustrate a comprehensive approach to customizing LLMs for EDA, showcasing the adaptability and potential of these models to revolutionize hardware design through nuanced, domain-specific enhancements.

Fine-tuning stands out as a common technique among projects including ChatEDA and VeriGen, where LLMs are tailored to understand and generate domain-specific language and processes. ChatEDA leverages fine-tuning with EDA tools to enable AutoMage in task planning and script generation, showcasing a model's adaptability to specific phases of the design flow. Similarly, VeriGen's approach to fine-tuning CodeGen-16B on a comprehensive Verilog corpus exemplifies how domain-specific knowledge can be embedded into an LLM, enhancing its ability to produce syntactically correct and functionally accurate HDL code.

ChipNeMo's use of domain-adaptive pretraining and supervised fine-tuning, coupled with custom tokenizers and retrieval models, demonstrates another layer of customization. By adapting LLaMA2 to the intricacies of chip design, the project significantly enhances model performance in EDA tasks, achieving a notable reduction in model size without sacrificing efficiency. This approach underscores the importance of domain-specific adaptation, allowing the model to navigate the unique challenges of EDA with improved accuracy and relevance.

Moreover, the prompt engineering used in RTLLM and the in-context learning approach of ChipGPT and GPT4AIGChip reveal the other innovative directions for customization. RTLLM's benchmarking and prompt engineering technique, focusing on syntax correctness, functionality, and design quality, highlight how nuanced prompt design can significantly enhance an LLM's output. Meanwhile, ChipGPT and GPT4AIGChip employ in-context learning to navigate design tasks, demonstrating the potential of leveraging existing model capabilities in novel ways to address complex hardware design challenges.

Autonomous agent framework is a promising direction to let general-purpose LLM adapt to hardware engineering problems. 
RTLFixer let LLM to read the compilation result of its generated code to fix the syntax error of RTL code. VerilogCoder designs a multi-agent flow by using a planner and integrating a waveform tracing tool to improve the functionality of generated RTL code. DRC-Coder lets LLM cooperating with vision language models to solve design rule coding with the help of layout image interpretation.

In essence, the customization techniques deployed across these projects reveal a strategic approach to adapting LLMs to the EDA domain. These methods showcase the versatility and potential of LLMs to revolutionize the field of hardware design. The careful adaptation of models into their application domains not only enhances their performance in specific tasks but also broadens the scope of what can be achieved in EDA.

\subsection{Multi-Modal Feature Representation}

The representation of design features within LLMs is another crucial aspect, influencing the models' ability to accurately interpret and generate design-related data. 
Effective feature representation involves encoding both the syntactical structure of design languages and the semantic relationships between different design elements, enabling LLMs to generate coherent and functionally relevant EDA outputs. 
This focus on feature representation stems from the need to bridge the gap between natural language processing capabilities and the technical requirements of EDA tasks.

Pre-exsiting research works in Table~\ref{tab:hdl-gen-sft} have predominantly employed textual representations of hardware design features, such as code snippets, documentation, and specifications in natural language. These representations enable LLMs to generate and optimize HDL code by understanding the syntax and semantics of the design languages. However, this approach mainly captures the linear, textual aspects of hardware designs, potentially overlooking the intricate relationships and dependencies inherent in hardware components and their connections.

Building upon existing research efforts, there is a significant opportunity to broaden the scope of feature representations. 
These conventional methods have laid a solid foundation, yet the complexity and multi-faceted nature of hardware design and verification suggest that leveraging multi-modality could unlock deeper insights and efficiencies. 
Multi-modal training is a prominent method for constructing the language-image model~\cite{radford2021learning}.
As we pivot towards the future, considering more diverse and intricate forms of data representation becomes crucial for advancing LLM capabilities in EDA.

Future directions for feature representations in LLM applications for EDA might encompass:
\begin{itemize}
\item \textbf{Graph-Based Circuit Features}: This direction involves encapsulating the hierarchical and interconnected structure of hardware components in a graph format. Such an approach allows for a more natural representation of relationships and dependencies in circuits, potentially enhancing LLMs' ability to navigate and optimize complex designs through graph neural networks.
Using GNN as an encoder for LLM~\cite{he2024gretrieverretrievalaugmentedgenerationtextual, chai2023graphllmboostinggraphreasoning} is a promising direction to enhance the LLM's graph understanding.


\item \textbf{Image-Based Circuit Features}: Another promising area is the use of image-based features, such as design rule description, layout-based images that capture cell density, pin density, and other spatial characteristics relevant to the backend implementation stage. This modality can provide LLMs with a spatial understanding of circuit layouts, enabling optimizations and analyses that were previously out of reach with purely text-based inputs.
DRC-Coder~\cite{chang2025drccoder} uses an multi-modal LLM to interpret the design rule description and layout visualization for benefiting the design rule understanding.

\item \textbf{Novel Text Representations}: Despite significant advancements in the use of textual data for EDA tasks, there remains untapped potential to explore novel text representations. These could include more detailed documentation of design rationale, constraints, and specifications that have not been extensively utilized in current LLM applications. Such enhancements in text representations could offer new dimensions of context and specificity to LLMs, further refining their output and decision-making processes.
Some works~\cite{mu2024learning, tan2024lloco} trains LLM to distill long context into a soft prompt to effectively represent the original prompt’s knowledge. For EDA, we can also develop our own text encoder to encapsulate the knowledge in EDA documentations and extremely-long EDA tool log for future use.
\end{itemize}

By embracing a multi-modal approach that incorporates graph-based and image-based features alongside evolving text representations, LLMs can achieve a more holistic understanding of hardware design tasks. This comprehensive perspective is essential for tackling the intricate challenges of EDA, paving the way for more nuanced, accurate, and efficient design and verification processes. As research in this area progresses, the integration of diverse data modalities promises to revolutionize the capabilities of LLMs in EDA, moving towards a future where AI-driven solutions offer unprecedented support in the realm of hardware design.

\section{Outlook on LLMs for EDA}

\subsection{Academic Infrastructure}

In this subsection, we briefly introduce the academic infrastructure for LLM for EDA research, including datasets, LLM backbones, etc.
The academic infrastructure for LLM research in EDA has seen significant development, with a growing number of open-sourced models and benchmarks/datasets becoming available. These resources provide a foundation for ongoing research and experimentation, facilitating the development of more effective and efficient LLM applications in the EDA domain. Open-sourced models allow researchers to explore different architectures and training techniques, while benchmarks and datasets enable the evaluation of model performance on standardized tasks, promoting progress and innovation in the field.

Regarding dataset, \cite{liu2023chipnemo} uses data from NVBugs (NVIDIA’s internal bug database), bug summary, design source, documentation, verification.
LLaMA2 tokenizer is adapted and approximately 9K new tokens are added to improve tokenization efficiency.
\cite{he2023chateda} query GPT-4 to generate and collect instructions.
The core controller, AutoMage is further fine-tuned on these instructions.
\cite{fu2023gpt4aigchip} uses open-source HLS code snippets from GitHub and customized HLS templates with implementation instructions to fine-tune LLMs.

\subsection{Application Bottlenecks}



The integration of LLMs into the EDA offers transformative potential, aiming to streamline the complex processes of designing and verifying electronic systems.
Existing commercial LLM-based EDA tools like \cite{RapidGPT} have emerged and shown preliminary success.
However, deploying LLMs in this specialized field encounters several industrial application bottlenecks.

EDA involves designing proprietary and often highly confidential electronic components and systems.
LLMs require vast amounts of data for training to provide meaningful insights and automation capabilities.
Companies may be reluctant to share sensitive design data needed to train these models, fearing intellectual property theft or leakage of confidential information.

The EDA domain encompasses a wide range of highly specialized knowledge, including circuit design, semiconductor physics, and manufacturing processes.
Training LLMs to understand and generate useful outputs in this context demands not only vast amounts of data but also highly curated and domain-specific datasets.
Achieving a level of understanding and output generation that meets the precision and accuracy requirements of the industry is a significant challenge.

EDA relies on a sophisticated ecosystem of software tools and workflows that have been developed and optimized over decades.
Integrating LLMs into this ecosystem without disrupting existing workflows or requiring significant changes in toolchains presents a logistical and technical challenge.
The models must be adaptable and flexible enough to work within the constraints of legacy systems and software.

The computational demands of both training and running LLMs can be significant.
EDA tasks often require real-time or near-real-time analysis and feedback, putting further pressure on computational resources.
Balancing the computational load of LLMs with the need for efficient processing times is crucial, especially for complex designs and simulations.

For EDA applications, where the cost of errors can be extremely high, ensuring the accuracy and reliability of LLM outputs is paramount.
Building trust in LLM-generated solutions among engineers and designers, who are accustomed to relying on deterministic tools and their own expertise, is a challenge.
Ensuring the models are robust, reliable, and capable of handling edge cases without introducing errors is critical for their acceptance.

The semiconductor industry is governed by strict standards and regulations to ensure the reliability and safety of electronic components and systems.
LLMs must be able to design and verify components that meet these stringent requirements.
Developing models that consistently adhere to these standards and ensuring that their outputs can be verified and validated against regulatory requirements adds another layer of complexity.

\subsection{Ethics, Security, and Efficiency}

The use of LLMs in all fields raises a number of concerns regarding data privacy, model interpretability, IP protection, computational constraints, and energy demand; in this section, we will briefly discuss these concerns.
To the best of our knowledge, there is currently no public work directly relating such concerns to using LLMs for EDA so we will first cover them from a general LLM perspective before discussing how they can relate to the EDA field and potential future research areas.

LLMs are often trained on large, internet-scale datasets that may include personal or confidential data which may be inadvertently exposed during inference.
Techniques such as differential privacy, federated learning, and homomorphic encryption have been proposed to mitigate these risks by preventing models from memorizing and revealing sensitive data \cite{feretzakis2024privacy}.
In EDA, quality designs are highly valuable, propriety information while also being the best resources for training models.
This makes it imperative that techniques are used that allow a model to learn from proprietary data while training without the possibility that it can be leaked during inference.

The "black box" nature of LLMs makes it inherently difficult to understand any reasoning present in their decision-making process.
As outlined by \cite{singh2024interpretability}, finding ways to create interactive natural language explanations alongside LLM output is an emerging and likely highly impactful field.
In EDA, understanding the rationale behind design decisions is crucial, and improved understanding will facilitate easier downstream tasks like debugging, verification, and validation.

Additionally, the deployment of LLMs in any field requires significant computational power leading to high energy consumption.
Because training happens only once and inference is continuous, we will consider inference-based energy consumption.
For example, a single text-based prompt and response query to LLaMA3-70B uses $2.26 \times 10^{-3} \ \text{kWh}$ of energy \cite{husom2024energy} which when considering the highly iterative process of designing with an LLM is significant.
Furthermore, electronic design with LLMs requires iterative prompting across many levels, leading to increased energy consumption.
This amounts to a significant carbon footprint in the majority of situations where the inference is not fully powered by renewable energy sources.
All of this points to a need for more efficient hardware to run LLMs on, as well as model optimization techniques to reduce the compute required for the same quality of results.

All of these topics are ongoing, popular research fields, and the EDA community will almost certainly benefit from advancements made by the greater LLM community.
However, there is an opportunity for impactful work to be done directly relating these concerns to EDA.
This could include showing that it is possible to extract full designs present in its training data from a model and adapting defensive techniques to prohibit such action or exploring LLM explainability directly in circuit design choices.


\bibliographystyle{ACM-Reference-Format}
\bibliography{ref}










\end{document}